%% file: main.tex
\documentclass[authoryear, twocolumn, a4paper]{article}

\usepackage{amssymb}
\usepackage{amsmath}
\usepackage{abstract} 
\usepackage{titlesec} 
\titleformat{\section}[block]{\large\scshape\centering}{\thesection.}{1em}{} 
\titleformat{\subsection}[block]{\large}{\thesubsection.}{1em}{} 

\usepackage{hyperref}
\usepackage{subfig}
\usepackage{rotating}
\usepackage{slashbox}
\usepackage{cleveref}
\usepackage{placeins}
\usepackage{natbib}
\usepackage{booktabs}

\usepackage{authblk}

\newcommand{\databaseurl}{\url{https://vision.eng.au.dk/plant-seedlings-dataset}}

\title{More than one Author with different Affiliations}
\author[1]{Thomas Mosgaard Giselsson\thanks{tgi@mmmi.sdu.dk}}
\author[2]{Rasmus Nyholm J\o rgensen\thanks{rnj@eng.au.dk}}
\author[3]{Peter Kryger Jensen\thanks{pkj@agro.au.dk}}
\author[2]{Mads Dyrmann\thanks{madsdyrmann@eng.au.dk}}
\author[1]{Henrik Skov Midtiby\thanks{hemi@mmmi.sdu.dk}}
\affil[1]{University of Southern Denmark, The Maersk Mc-Kinney Moller Institute}
\affil[2]{Aarhus University, Department of Engineering - Signal Processing}
\affil[3]{Aarhus University, Department of Agroecology - Crop Health}

\title{A Public Image Database for Benchmark of Plant Seedling Classification Algorithms}
\begin{document}
\maketitle
\begin{abstract}
\input{src/abstract}
\end{abstract}


\input{src/intro}\label{sec:intro}
\input{src/databaseIntro}\label{sec:dbintro}
\input{src/imagerecording}\label{sec:recording}
\input{src/segmentation}\label{sec:segmentation}
\input{src/conclusion}\label{sec:conclusion}
\input{src/appendices}\label{sec:appendices}
\input{src/acknowledgement}

\bibliographystyle{apalike} 
\bibliography{bibliography/noURLcitationKeyBib}





\end{document}

%% file: src/abstract.tex
A database of images of approximately 960 unique plants belonging to 12 species at several growth stages is made publicly available.
It comprises annotated RGB images with a physical resolution of roughly 10 pixels per mm.
To standardise the evaluation of classification results obtained with the database, a benchmark based on $f_{1}$ scores is proposed.
The dataset is available at \databaseurl.

%% file: src/intro.tex
\section{Introduction}
For several decades, researchers have worked on systems aimed at performing site-specific weed control.
The approaches used range from weed maps constructed using coarse remote sensing data to real-time precision spraying using ground-based platforms equipped with high-resolution imagery sensors.
Common to all approaches is the goal of detecting weeds - either in patches or as single plants.
Although some systems are commercially available, a true commercial breakthrough of such systems is still to come despite the construction of several prototypes and case studies showing promising results.
The reason may be that a general approach enabling robust classification despite varying conditions and species compositions is yet to be discovered.
One might ask why this task still poses a problem: botanists have been dealing with species categorization for centuries and substantial progress is reported in the field of content-based image retrieval and analysis of images and video.
What makes this problem so hard? The present authors believe that one problem is a lack of benchmark databases.
Several studies on species recognition contain a description of preprocessing steps such as image acquisition, segmentation and annotation which suggest that researchers have spent time on these topics although each of these tasks is an area of its own.
To support and encourage the development of species recognition techniques for the agricultural industry, this paper presents a database that is freely available to researchers and which enables them to jump directly to the task of object analysis, species recognition or plant appearance analysis.
Furthermore, a performance benchmark for classification is proposed, so that using this database will permit easy replication of research results and easy comparison of algorithm performance.

Other databases exist but vary in their availability and content which prevent them from being usable for specie recognition approaches aiming at site-specific weed control. In \cite{Agarwal2006} it is argued that botanists' databases such as the Smithsonian database and other world herbaria databases do not have representative samples that capture within-species variance.

In \cite{Belhumeur2008} a system that can help botanists identify plant specie from leaf shapes is described.
They state that \citet{Soderkvist2001} describes a publicly available database, but the present authors have had no luck in locating that database.
 Recently also \cite{Meyer2011} has stated that many databases exist, but their usability for developing machine vision systems is yet to be determined.

Developing a generic specie recognition system fails because of the lack of knowledge or ability to cope with a huge variation in specie appearances. This work aims to help researchers acquire insight into the normal variation of some of the most common weed species in Danish agriculture.
It is believed that if a classification approach is able to handle this data then it is likely to possess the ability to cope with high intra-class variations and be a step in the right direction towards automatic specie identification, usable for example for site-specific weed management.
The authors hope that others will try their methods on this dataset and use it for benchmarking against alternative approaches.

The data described in this paper is aimed at ground-based weed or specie spotting.
Respected researchers in the domain have argued that the most promising approach for site-specific weed control is currently a ground-based computer vision system \citep{Gerhards2010}, since approaches such as remote sensing need to address a myriad of problems in order to be robust and reliable - many of which (such as solar angle and cloud cover) are beyond the control of the users \citep{Thorp2004}.

%% file: src/databaseIntro.tex
\subsection{Image recording}
Many research groups around the world work on plant recognition. Often some kind of image database of specie samples is used for evaluating their approach. Creating a sample database requires a lot of time and planning. Data is recorded with various equipment ranging from off-the-shelf commercially available cameras to specially constructed sensors for a specific data acquisition task. This means that data recorded by different researchers are of different quality, recorded by different sensor types and of different quantities not to mention of different species. The result is that data is diverse and can be sparse and often collected for very specific research.  Furthermore data is not always made available to other researchers and comparison of different methods by comparing results from published papers is therefore hard or impossible.
The author is not aware of any publicly available database of plant seedling images comprising the number of samples necessary to assess performance of plant specie recognition methods targeted at site-specific weed control.
Having a publicly available database would enable researchers to test their methods against a common dataset, enable comparison and hopefully encourage research because of a better overview of performance of different methods.
This paper presents such  a public database of plant seedling images.

To build such a database, the following considerations need to be addressed: which species should be used; at what time(s) should recording take place; how many samples should there be of each specie; what equipment should be used for image acquisition; and how should illumination be handled.
Decisions taken to cope with these choices constrain the generality of the database, but are necessary to make the task achievable.
As long at the acquisition process and equipment used are thoroughly documented, the database is expected to be of high value to other researchers.

Since it is not possible to include too many species, only a subset of high importance to the Danish agricultural industry are chosen.
Styrofoam boxes are used to grow samples.
Each box contains on average 25 samples.
A total of 56 boxes are used, with only one specie sown in each box.
It is believed that 80 samples of individual species at the same growth stage is sufficient to capture the main variations within a specie, so 4 boxes of each specie are sown so as to allow for 20\,\% germination failure.

Images are recorded multiple times over a 20 day period at an interval of 2 to 3 days, starting a few days after emergence.
(The database is primary targeted to research that tries to identify plant species at an early growth stage, so  as to allow farmers to conduct weeding before weeds start competing with crops for nutrition.) A dSLR camera (Canon 600D) is used with a fixed 50mm lens for recording RGB images.
The camera is placed approximately 110 - 115 cm above the soil surface, and has a native resolution of 5184 x 3456 px which result in images having a physical resolution of approximately 10 pixels per millimeter.

%% file: src/imagerecording.tex
\subsection{Specie selection}

14 important weed and crop plant species appearing in Danish arable fields have been chosen for this work.
These species are a subset of plants that any site-specific weed control system operating in Denmark will need to handle.
The specie list covers monocotyledon and dicotyledon crop and weed species and is given in Table~\ref{tab:species}. Unfortunately two species did not germinate at all(Redshank, Field Pansy) resulting in 12 species being recorded.


\begin {table*}
\centering
\caption {Table of species included in the image database. The ID referrers to styrofoam box ID and folder names in the database. IDs marked with * did, unfortunately, not germinate} \label{tab:species}
\begin{tabular}{ l  l  l  l }
\toprule
\textbf{Danish} &\textbf{English} &\textbf{Latin} &\textbf{ID} \\
\midrule
Majs &Maize &Zea mays L. &1-4 \\
Vinterhvede &Common wheat &Tricicum aestivum L. &5-8 \\
Sukkerroe &Sugar beet &Beta vulgaris var. altissima &9-12 \\
Lugtl\o s kamille &Scentless Mayweed &Matricaria perforata M\'{e}rat &13-16 \\
Fuglegr\ae s &Common Chickweed &Stellaria media &17-20 \\
Hyrdetaske &Shepherd's Purse &Capsella bursa-pastoris &21-24 \\
Burresnerre &Cleavers &Galium aparine L. &25-28 \\
Fersken pileurt &Redshank &Polygonum persicaria L. &29-32* \\
Agersennep &Charlock &Sinapis arvensis L. &33-36 \\
Hvidmelet g\aa sefod &Fat Hen &Chenopodium album L. &37-40 \\
Liden storken\ae b &Small-flowered Cranesbill &Geranium pusillum &41-44 \\
Agerstedmoder &Field Pansy &Viola arvensis &45-48* \\
Agerr\ae vehale &Black-grass &Alopecurus myosuroides &49-52 \\
Vindaks &Loose Silky-bent &Apera spica-venti &53-56 \\
\bottomrule
\end{tabular}
\end{table*}

\subsection{Image recording}
Weed control needs to be carried out as early as possible after crop germination to avoid the adverse effects on crops of competition with weeds.
Broadcast spraying with selective herbicides is often performed before 20 days after crop germination if conditions allows for it.
Based on this timing, researchers try to accomplish weed assessment or control within the same time frame.
In \cite{Downey2004} they collect images 10 days after planting, and \cite{Woebbecke1995a}  argues that after between 14 - 23 days of plant development is a favorable time for post-emergence weed control with respect to plant appearance and weed control needs.
At an early growth stage the common problem of overlapping plant leaves is also minimal and makes the resulting image material more manageable.
Non-overlapping plants enable analysis of single plants for specie categorization --- the main obstacle for commercializing automatic site-specific weed control systems, as argued by \cite{Slaughter2008} in their comprehensive and often cited review paper.

\subsection{Acquisition setup}
Plants are sown in styrofoam boxes of dimension 270 by 210 mm and nursed and grown at AU-AgroEcology research facility in Flakkebjerg\footnote{Department of Agroecology - Crop Health, Fors\o gsvej~1, 4200 Slagelse, Denmark}.
The plants grow in soil, the surface of which is covered with small stones.
Earlier experiments showed that bare soil in indoor moist conditions tends to develop a green moss layer that can be a source of error for pixel-based segmentation algorithms.


\subsection{Tray and specie tracking}
Each box is sown with a single specie and is labeled with a number, both as a numeral and as a bar code.
Having only a single specie per box eases tracking of species.
For every specie 4 boxes are sown,  ensuring that unforeseen conditions affecting a random box will not affect all samples from a single specie and thereby be misleading.
The repeated boxes also ensure the number of samples aimed at.
Table \ref{tab:species} lists the IDs for each sample.


\subsection{Recording considerations}
\begin{figure*}
\centering
\subfloat[Camera rig]{%
\includegraphics[height=5cm]{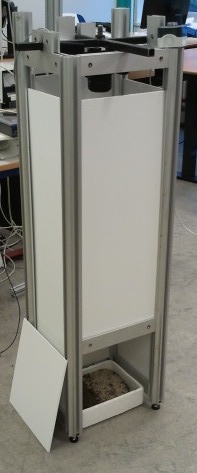}
\label{fig:rig} %
}
\subfloat[Inside of camera rig]{ %
\includegraphics[height=5cm]{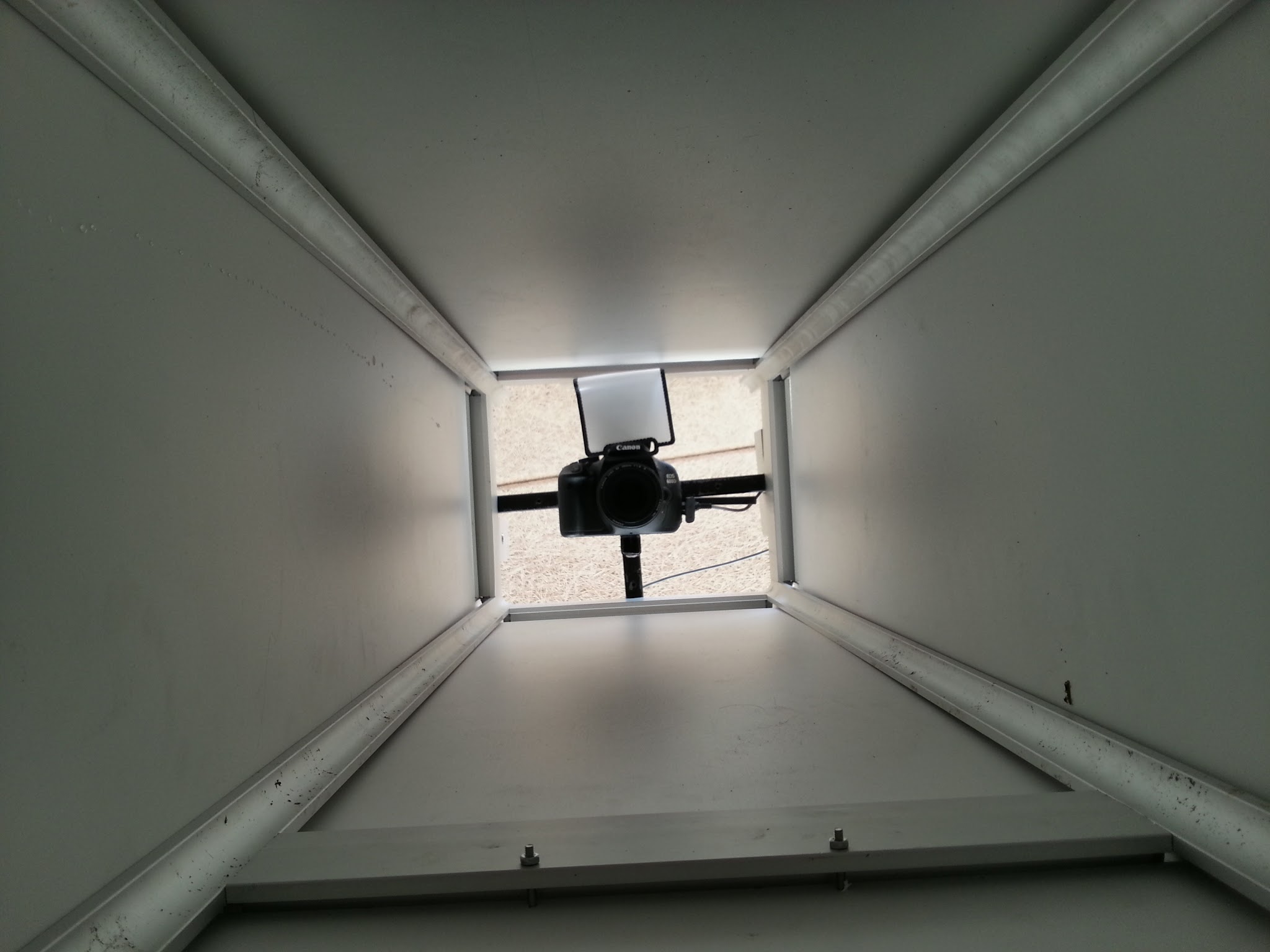}
\label{fig:inside}%
}
\subfloat[Styrofoam boxes]{ %
\includegraphics[width=5cm, angle= 90]{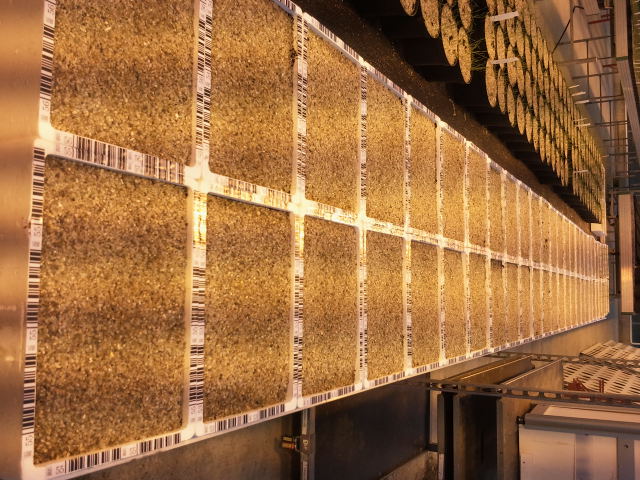}
\label{fig:overview} %
}

\caption{Recording rig and the used styrofoam boxes. \protect\subref{fig:rig} Camera rig constructed from ITEM profiles and white rigid sheets - camera is fixed in the top facing down. \protect\subref{fig:inside} Inside of camera rig. The build in blitz of the camera were used with a diffuser infront. \protect\subref{fig:overview} Styrofoam boxes sown with plant and covered with small stones}
\label{fig:recordingAndRig}
\end{figure*}

To acquire images a custom camera rig was constructed. The final rig can be seen in figure \ref{fig:rig}. The following text will list and explain considerations related to the rig construction.

\paragraph{Rig}
A rig constructed of ITEM™ sections was used to hold the camera, ensuring a fixed camera height with respect to the soil surface of between 1100 and 1150 mm.
Spaces between the ITEM™ structure were filled with a rigid sheet having a rough white surface, ensuring scattering of light (see figure \ref{fig:rig}).

\paragraph{Light}
To ensure even and comparable light conditions, the camera's built-in flash was used.
A diffusion screen was put in front of the flash as a measure to avoid hard shadows.
The sides of the rig also worked to minimize point-like light source artifacts. This is depicted in figure \ref{fig:inside}.

\paragraph{Vibration}
To avoid any sort of motion blur, an external trigger was used with a small delay of 2 seconds between triggering the camera and the acquisition of the image by the camera so that any vibration had a change to diminish.
\paragraph{Lens and focal length}
The camera was equipped with a fixed CANON lens EF 50\,mm 1:1.
8 II (serial no. 8791016624) set to Auto Focus.
Using the constructed rig resulted in the camera being approximately 1.1 to 1.15 meters above the soil surface.
With a fixed focal length of 50\,mm and a camera sensor size of 22.3 by 14.9\,mm, one can calculate the field of view in the horizontal and vertical directions.
These calculations are based on simple trigonometry \citep{Pentax1998} and result in equation \ref{eq:fov}.

\begin{align}
\label{eq:fov}
\theta = 2\cdot arctan \left( \frac{D}{2f} \right)
\end{align}
where D is the sensor size, f is the camera focal length and $\theta$ is the viewing angle.
The depth of field can be calculated using the previously stated camera and lens parameters together with the COC value (circle of least confusion).
For the Canon 600D camera, the COC is 0.019\,mm \footnote{\url{http://www.dofmaster.com/digital_coc.html}}.
The near ($T_{near}$) and far ( $T_{far}$) focus limits can be calculated by equation \cref{eq:h,eq:tnear,eq:tfar}
\begin{align}
H&=\frac{f^2}{coc \cdot F} \label{eq:h} \\
T_{near}&=\frac{D_{obj} \cdot (H+f)}{H+D_{obj}} \label{eq:tnear} \\
T_{far}&=\frac{D_{obj} \cdot (H-f)}{H-D_{obj}} \label{eq:tfar}
\end{align}
where H is the hyper focal distance, $D_{obj}$ is the distance to the object in focus and F is the aperture f-number.
Calculating depth of field for objects at a distance of 1100.0 mm results in a range from 1041.17 to 1166.26 mm spanning 125.09\,mm. Likewise when calculating depth of field for objects at a distance of 1150.0 mm, the range is from 1085.73 to 1222.78\,mm spanning 137.05 mm. 
Plant were foreseen to grow not higher than 100\,mm during the recording period, meaning that object height would constitute less that 10\,\% of the camera-to-object distance and be within acceptable distance with respect to the depth of field.

\paragraph{Resolution}
Images were acquired with a high resolution DSLR camera giving 5184 $\times$ 3456\,px.
This means that the resolution will probably be higher than any practical commercial system, making the database ideal for testing;  the premise being that algorithms not working on detailed and high resolution images will also not work on images of lower quality.
Calculating the approximate pixel resolution for the object distance limits mentioned earlier results in 10.6 pixels per mm when object distance is 1.1 meters and 10.1 pixels per mm when object distance is 1.15 meters.

\subsection{Recording procedure}
The camera was set to an ISO number of 100, a shutter speed of 1/60 second and an aperture of f7.1.
After germination each box was photographed every 2 or 3 days for a 3 week period or until the box just consisted of a green cover.
Each box was individually manually placed in the camera rig  each time.

\begin{figure*}
\centering
\includegraphics[width=8cm]{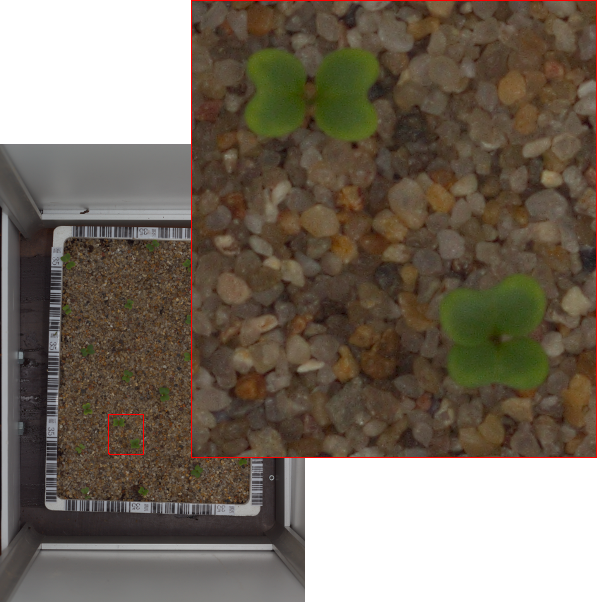}
\caption[Example]{Example image showing box 35 containing Charlock plants. The extracted close-up gives a feel of detail degree}
\label{fig:sample}
\end{figure*}

\paragraph{Example data}
Figure \ref{fig:sample} shows a random sample extracted from the recorded image set, together with a close-up of the marked rectangle.

\subsection{Proposed benchmark measure}\label{subsubsec:benchmark}
With common public data material as presented in this paper, researchers are one step closer to being able to compare results.
Yet another step would be to agree on a performance measure.
We propose to use measures derived from classification results achieved by stratified cross-validation of any preferred classification algorithm.
The proposed procedure is as follows.
The data to be used is shuffled and divided into a number of disjoint sets (folds).
The classifier under test is trained with data from all but one of these folds, and its performance after training is evaluated with the remaining one.
Performance is measured using weighted averages of $f_1$ scores for each fold, where the weighting should be according to class sample size.
This procedure results in one performance measure for the classifier per fold.
Now assume that the set of weighted average $f_1$ score, $S_{f}$, constitutes a set of samples from a population of possible performance results for the classifier under test.
(The $f_1$ score is a commonly used measure \citep{Lu2009} and is invariant against sample size and the ability of a classifier to recognize true negatives \citep{Sokolova2009}.) This measured performance is a random variable since the data were selected randomly by shuffling.
We perform bootstrapping of the set $S_{f}$, to generate alternative possible sample sets from the same population, and generate 1000 bootstrap sets.
Calculating the average of each bootstrap set produces new samples of the underlying population of performance measurements.
Generating these samples enables the calculation of confidence intervals.
One simple way of doing this would be to sort the bootstrap set averages into ascending order, $L_{avg}$,  and extract, for the 95\% confidence interval, values at the indices $0.05*0.5*1000 = 25$ and $1000 - 0.05*0.5*1000 = 975$. The proposed measures for reporting classifier performance are then the fold average of weighted average $f_1$-score and its associated confidence interval achieved by bootstrapping.
We recommend using 10 fold cross validation.

Equation \ref{eq:pc} to \ref{eq:s} describe the calculated measures.
\begin{align}
P_c&=\frac{TP_c}{TP_c+FP_c} \label{eq:pc}\\
R_c&=\frac{TP_c}{TP_c+FN_c} \label{eq:rc}\\
f_{1,c}&=2\frac{P_c \cdot R_c}{P_c + R_c} \label{eq:f}\\
avg_{weighted}(f_1)&=\sum_{c=1}^{C}{\frac{N_c}{N} \cdot f_{1,c}} \label{eq:avgf}\\
S&=\frac{1}{N_{fold}}\sum_{i=1}^{N_{fold}}avg_{weighted}(f_1)_i \label{eq:s} 
\end{align}
where $TP_c$, $FP_c$ and $FN_c$ denotes True positives, False positives, and False negatives for class c respectively. $P_c$ is class specific precision and $R_c$ is class specific recall. $N$ denotes the total number of samples and $N_c$ the number of samples of class c and $C$ the total number of classes. $S$ is the mean of $avg_{weighted}(f_1)$ across all cross validation folds.
The $f_1$-score could be used for comparing both segmentation algorithms and specie classification algorithms.
Assessing segmentation algorithms will, however, be difficult since no ground truth is available.
For specie classification, the necessary annotation information is present and calculating $f_1$-scores is straightforward.
We therefore only propose a benchmark for classification here, though we plan to extend the database with annotations of ``correct'' segmentations which would facilitate the definition of a segmentation benchmark.

\subsection{Discussion --- The value of the database}

The authors regard the value of the database to be manifold.
With the data being publicly available, hopefully it will be used by researchers to test approaches for plant specie recognition at early growth stages.
 
The authors also believe that several themes not addressed in this paper will also benefit from the database.
From a botanist's point of view as well as that of computer vision engineers, this database could be used to gather insight into the morphological and structural development of plants during their early growth stage.
General object recognition researchers might benefit from working with data where the intra-class variance is large compared to the inter-class variance.

The image database is recorded with known limitations and trade-offs.
First of all, plants are grown indoors in a greenhouse with artificial light to supplement natural light.
The fact that data are recorded under laboratory conditions means that some aspects of variance will not be present and some morphological features might be different from outdoor-grown plants.
No insect or pest damage is present in the images and the somewhat unnatural light means that plants tend to stretch in an attempt to capture more light.
In essence growing plants indoors changes the phenotypic influence of the plant appearance \citep{Royer2005}.

Use of a high resolution dSLR camera resulted in a physical resolution of approximately 10 px/mm or 0.0001 m/px. This is comparable with resolutions reported by \cite{Thorp2004} where they state that resolutions as low as 0.0003 m/px are used. Some plant leaf features such as venation do require a very high resolution. In \cite{PLOTZE2009} they take taxonomy protocols as their starting point. This approach uses venation features that are not believed to be present in database samples due to resolution and plant stage.
Micro-resolution on the 0.1\,mm/px scale is not practically achievable on field machinery. To achieve image quality and resolution comparable to practical image recording devices, downsampling and addition of noise should be considered.

3D information would have increased the value of the database, especially with respect to monocotyledon plants that mainly grow upward in their early growth stages.

Besides using the database for benchmarking the authors also hope that researchers will make their algorithmic implementations available to others to ease further development and enable researchers to focus on specific aspects by using others' work for pre- or post-processing.

%% file: src/segmentation.tex
\section{Segmentation}

As an example of processing data from the database, we demonstrate a segmentation approach using a naive Bayes classifier.
Many researchers have investigated segmentation of vegetation images into vegetation and background.

\cite{Woebbecke1995259} reviewed and investigated different mappings of RGB colours to qualify how to achieve a best possible linear separation of green plant material and soil material.
The conclusion was that a linear combination of the form $2g-r-b$, with the chromatic values $r$, $g$ and $b$ or a modified hue value, offered the best separation capability between soil/residue and green plant material.

In \cite{Tian1998} they investigate the performance of an environmentally adaptive segmentation algorithm over a static segmentation algorithm. Their approach consists of a semi-supervised clustering approach that generated a dataset subsequently used for training a Bayesian classifier. They show that under changing lighting conditions their approach outperforms a static classifier.

\cite{Meyer2008a} also examined different mappings of the chromatic values of RGB images and included an investigation of the effect of varying background material. 
One conclusion of their work was that when images consisted of soil as background and green plant material as foreground, then ExG (Excessive Green) with an automatic Otsu threshold~\citep{2388173} and ExG-ExR with ExR being Excessive Red (boils down to G-R) with a static threshold of 0 performed equally well - both achieved a score of approximately 0.87 calculated by dividing the true positives by (true positives + false negatives + false~positives).

Approaches that make a linear transformation of sensor data are widely used. 
They often suggest a transformation which has proved to work well under certain circumstances.
Segmentation criteria such as Excess Green might give a visually pleasing result, but using approaches requiring a training step seems more well founded since they are inherently based on the apparent dataset.
For the demonstration of a segmentation approach in this paper a Bayesian classifier is trained and the result is visually inspected.

\subsection{Image segmentation example \mbox{using} Na\"ive Bayes}
\begin{figure*}
\centering
	\subfloat[Raw]{
		\includegraphics[width=4cm]{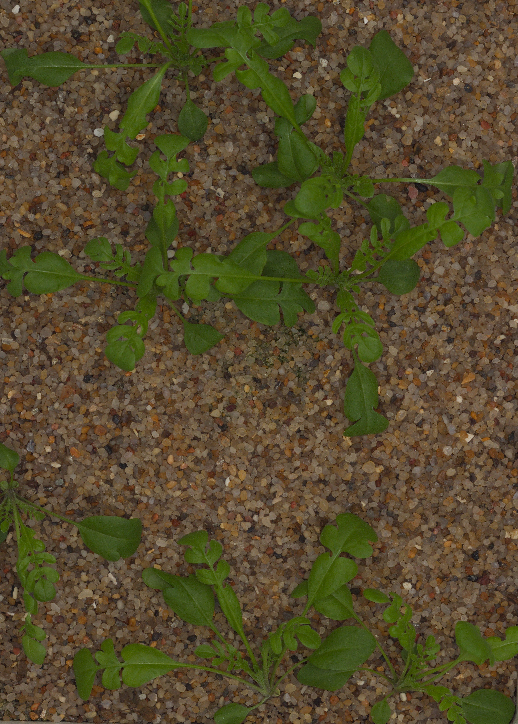}
		\label{subfig:input}
	}
	\subfloat[Uniform sampling]{
		\includegraphics[width=4.2cm]{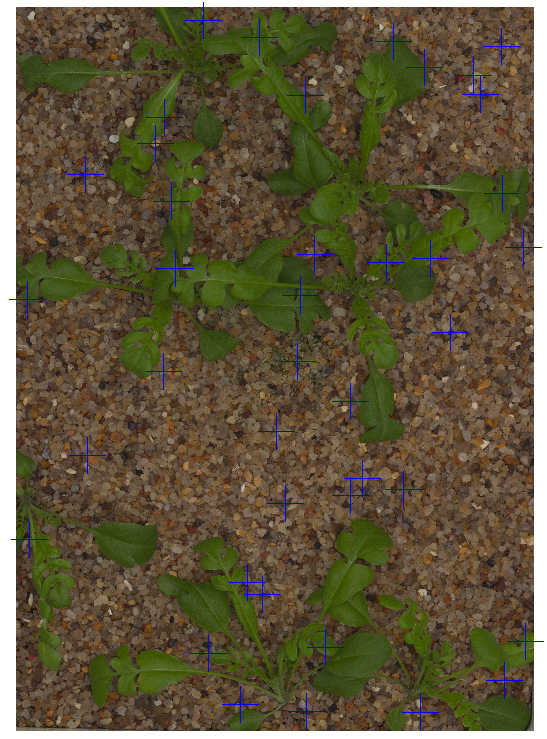}
		\label{fig:sampling}
	}
	\subfloat[Segmented]{
		\includegraphics[width=4cm]{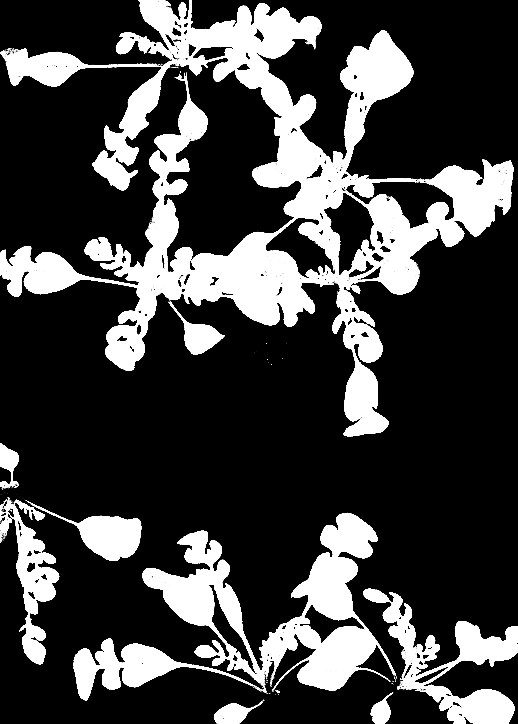}
		\label{subfig:posterior}
	}
\caption{Calculating the posterior probability of each pixel belonging to the foreground (plant material). Figure \protect\subref{subfig:input} shows the raw input image. Figure \protect\subref{fig:sampling} shows the uniform sampling from the image used for constructing a naive Bayes classifier. Figure \protect\subref{subfig:posterior} shows the posterior probability of a pixel being foreground where black and white is 0 and 1 probability respectively}
\end{figure*}

The following section will show the result of applying a naive Bayes segmentation approach to detect vegetation pixels in an image consisting of vegetation and small stones (see \ref{subfig:input}).
Since the image is a color image each pixel consists of a red, green and blue value, the classifier's input is a vector of size 3. The class, $C$ is encoded as being 1 for vegetation and 0 for background.
It is assumed that the distribution of color values is approximately gaussian \citep{Tian1998} so that $p(x|C=k)=\mathcal{N}(\mu_k,\Sigma)$ for $k=\{0,1\}$, where $\mu$ is a vector of size 3 of mean values and $\Sigma$ is a $3\times3$ covariance matrix. The variances of the two distributions are here assumed to be equal.
Given this model, one can write the likelihood function as a function depending on the unknown parameters and a sample set


\begin{equation}
\begin{aligned}
p(C,data| \mu_1,\mu_2,\Sigma, \theta)=&\prod_{n=1}^N \left(\right.[\theta \mathcal{N}(x_n|\mu_1,\Sigma)]^{t_n} \\
&\cdot [(1-\theta) \mathcal{N}(x_n|\mu_2,\Sigma)]^{1-t_n}\left.\right)
\end{aligned}
\end{equation}

Taking the log of this and maximizing the likelihood function with respect to each parameter in turn gives the maximum likelihood solutions for the parameters. The result becomes

\begin{equation}
\theta=\frac{N_1}{N}
\end{equation}
\begin{equation}
\mu_1=\frac{1}{N_1}\sum_{n=1}^Nt_n x_n
\end{equation}
\begin{equation}
\mu_2=\frac{1}{N_2}\sum_{n=1}^N(1-t_n) x_n
\end{equation}

\begin{equation}
\begin{aligned}
\Sigma = &\frac{1}{N}\sum_{n\in C_1}(x_n-\mu_1)(x_n-\mu_1)^T\\ 
+ &\frac{1}{N}\sum_{n\in C_2} (x_n-\mu_2)(x_n-\mu_2)^T 
\end{aligned}
\end{equation}

where $N_1$ is the number of samples from the vegetation (foreground) class and $N$ is the total number of samples.
Now the next step is to apply the algebra to an actual data set. To do this, some samples need to be collected.
This was done by sampling 40 times from an image from the database where each sample point was chosen by a uniform distribution covering the whole image.
The sampling can be seen in figure \ref{fig:sampling}.


Using the stated formula one can calculate the parameters. This resulted in the following values:

\begin{align}
\theta &= 0.3250\\
\mu_1 &= 	
\begin{pmatrix}
	0.2600\\
	0.3306\\
	0.0817
\end{pmatrix} \\
\mu_2 &=
\begin{pmatrix}
	0.3695\\
	0.2940\\
	0.2173
\end{pmatrix} \\
\Sigma &=
\begin{pmatrix}
0.0071 & 0.0055& 0.0036 \\
0.0055 & 0.0053& 0.0039 \\
0.0036 & 0.0039& 0.0043
\end{pmatrix}
\end{align}
Using these values the posterior distribution can be calculated for a random input vector. This has been done for all pixels in an image and the result can be seen in Figure~\ref{subfig:posterior}.

\subsubsection{Segmentation result and discussion}
\begin{figure*}
\centering
\includegraphics[width=8cm]{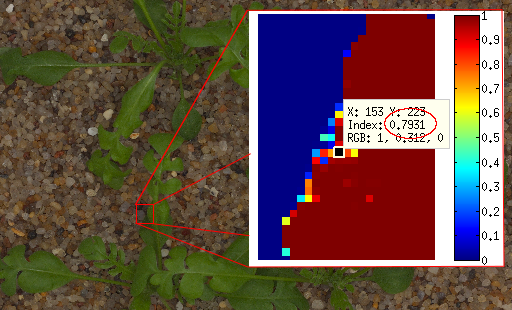}
\caption{Uncertainty in the edge between foreground and background. The color coding correspond to blue being 0 \% and red being 100 \%}
\label{fig:edge}
\end{figure*}

The result is a functional segmentation method that emphasizes vegetation pixels relative to the background. On the edge between foreground and background the segmentation is less certain which can be seen on figure \ref{fig:edge}. The constructed classifier is based on samples and prior distributions collected from a single image which means that the classifier is only usable on images with similar foreground and background distributions. 
Another segmentation approaches using the presented image database are described in \cite{Dyrmann2013}, from which the individual plants are also provided at \databaseurl. Segmented samples from \cite{Dyrmann2013} are shown in \ref{sec:samplePlants}.

As stated by other researchers, segmentation performance is often assessed by visual inspection and this is also the case in this paper.
Segmentation is, in general, difficult to assess for several reasons: segmentation approaches often have difficulties only in the boundary region between foreground and background.
This region is also difficult, if not impossible, for humans to annotate correctly and this makes it very hard to achieve correct ground truth with real plant data.
This is because of the physical construction of cameras - each pixel measures an average over some area, meaning that no sharp edge between foreground and background exists.
Lens aberration can also introduce severe noise in the pixels in the close vicinity of the true edge.

Note that the current version of the database does not contain ground truth segmentation labeling, so it is not straightforward to define a segmentation benchmark using the database.


%% file: src/conclusion.tex
\section{Conclusion}
A public database of images of 12 species common in Danish agriculture is presented.
Plants have been grown indoors in styrofoam boxes and recorded multiple times over a 20 days period.
A total of 407 images have been recorded and are made available on the internet (\databaseurl)) at the time of publication. Each image is named with an ID that relates the image to a single specie. The image database is aimed at researchers working with specie recognition;  as well as describing the data, this paper also suggests a benchmark measure to researchers to be able to compare classification results.

For demonstration purpose a segmentation process using a naive Bayes classifier is documented.

%% file: src/appendices.tex

\appendix
\section{Sample plants}\label{sec:samplePlants}
\begin{figure}
\centering
	\subfloat[Cleavers]{
		\includegraphics[width=0.3\textwidth]{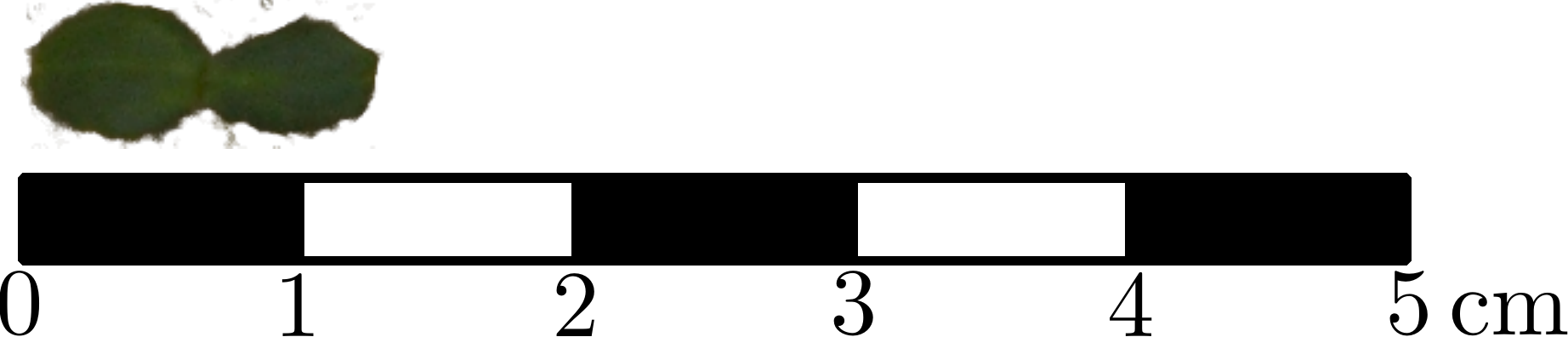}
	}
	\hfill
	\subfloat[Loose Silky-bent]{
		\includegraphics[width=0.3\textwidth]{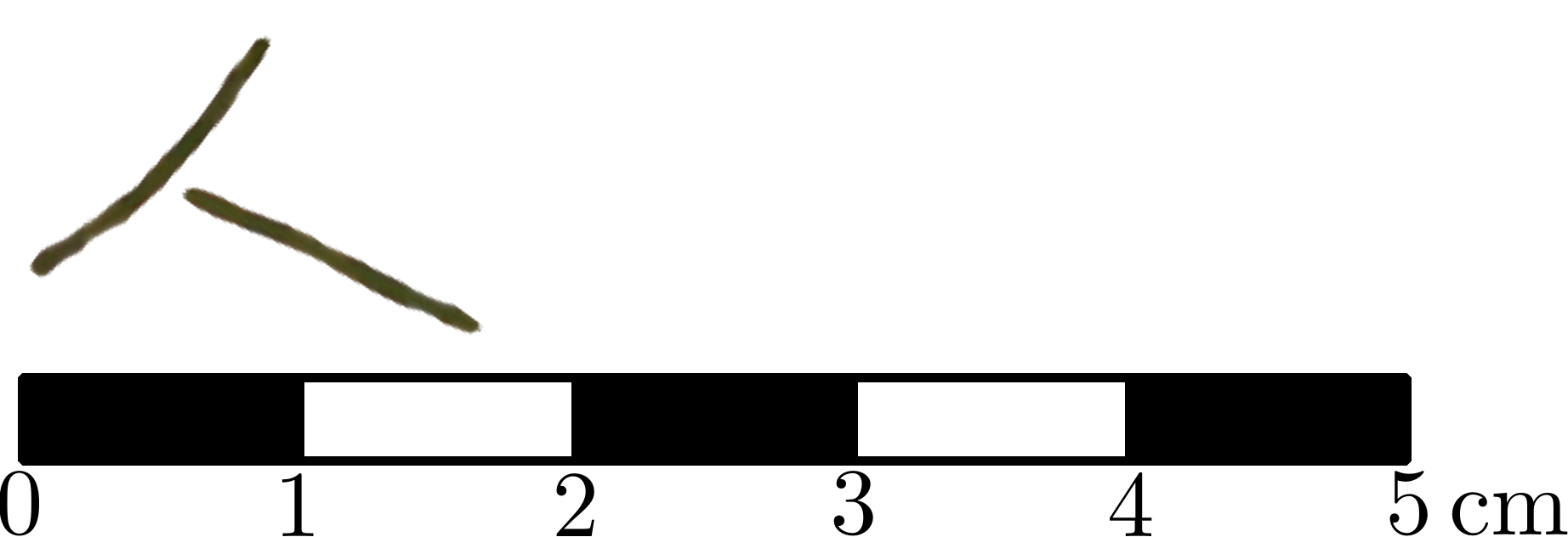}
	}
	\hfill
	\subfloat[Sugar beet]{
		\includegraphics[width=0.3\textwidth]{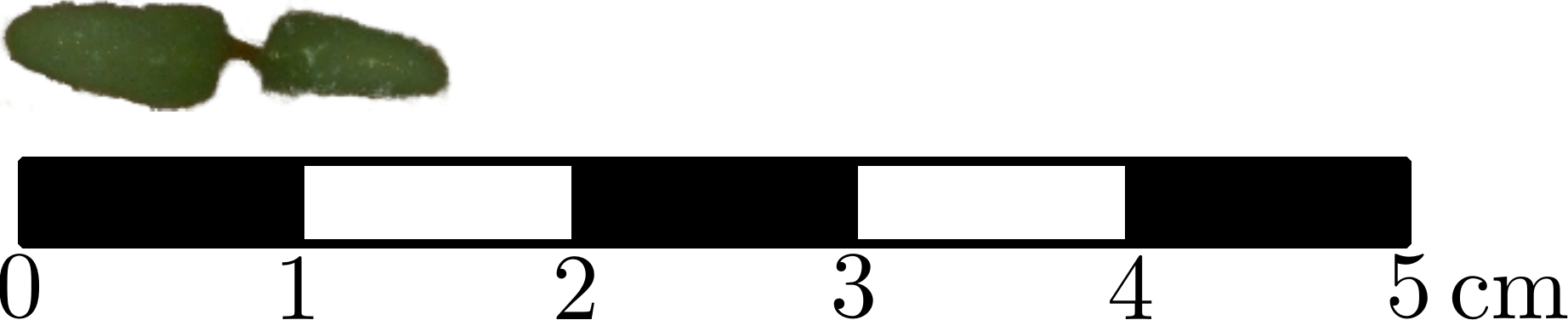}
	}
	\hfill
	\subfloat[Charlock]{
		\includegraphics[width=0.3\textwidth]{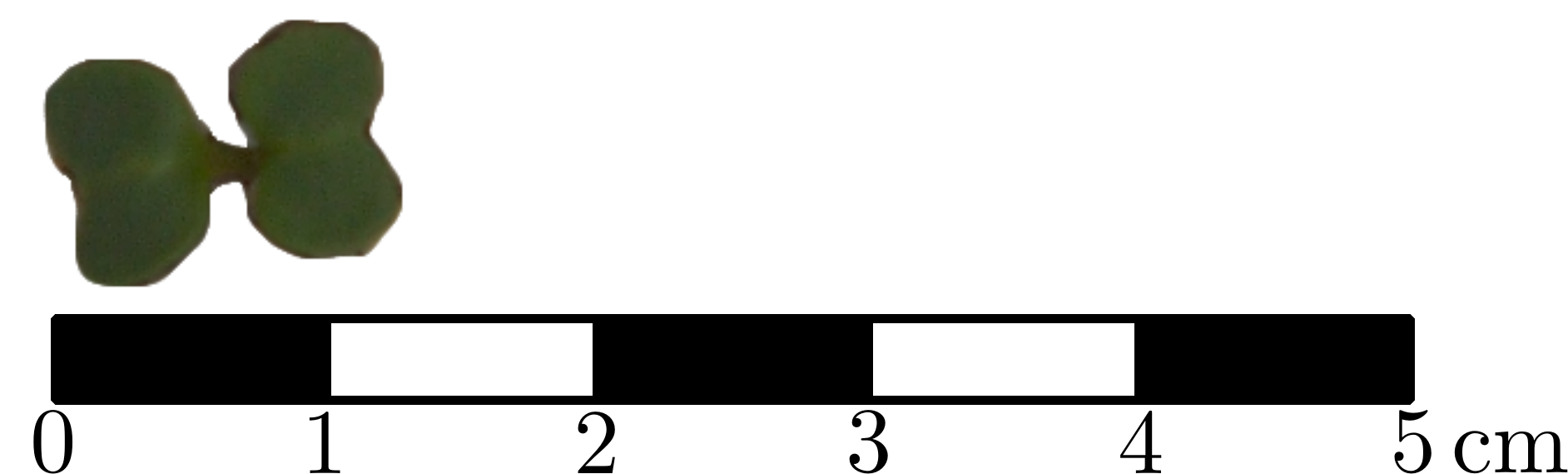}
	}
    \hfill
	\subfloat[Scentless Mayweed]{
		\includegraphics[width=0.3\textwidth]{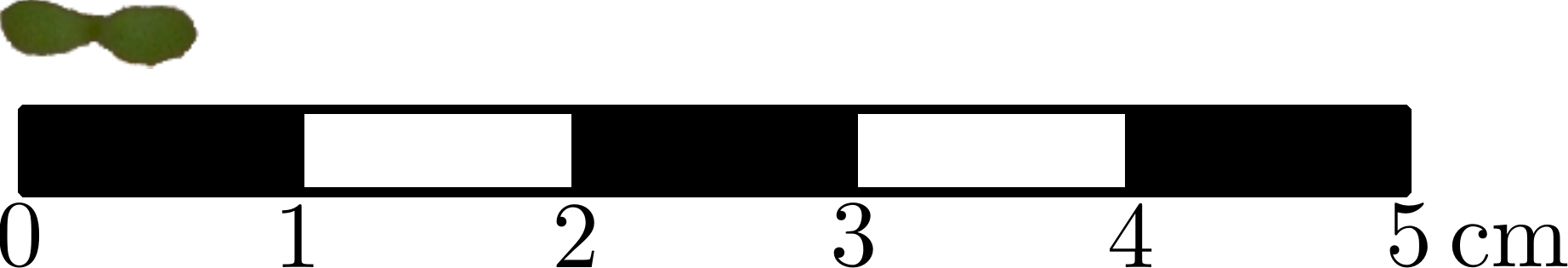}
	}
	\hfill
	\subfloat[Maize]{
		\includegraphics[width=0.3\textwidth]{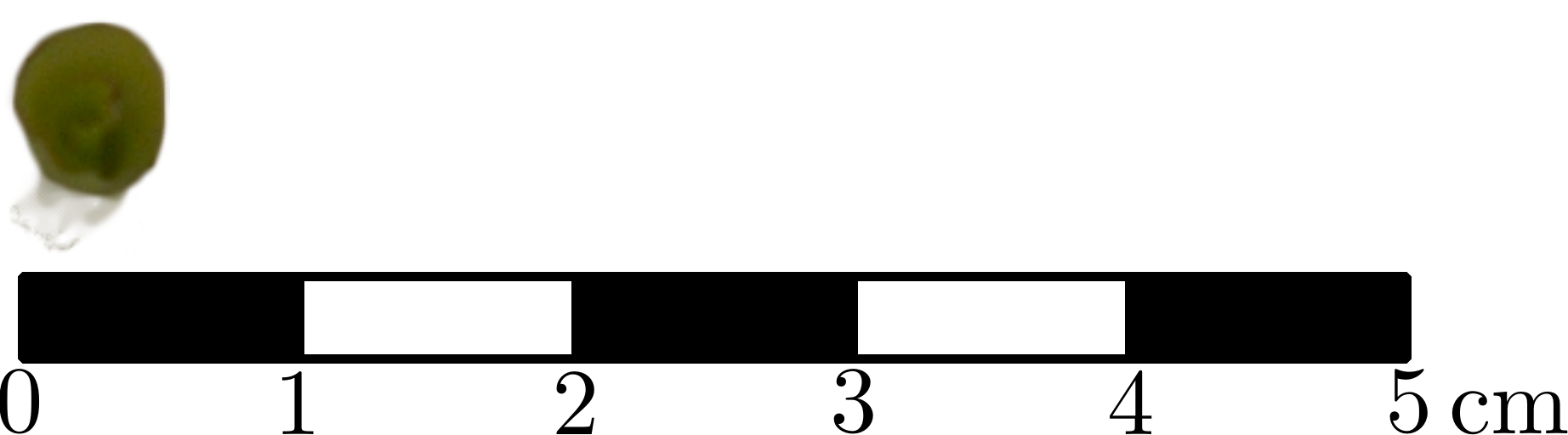}
	}
    \hfill
	\subfloat[Common Chickweed]{
		\includegraphics[width=0.3\textwidth]{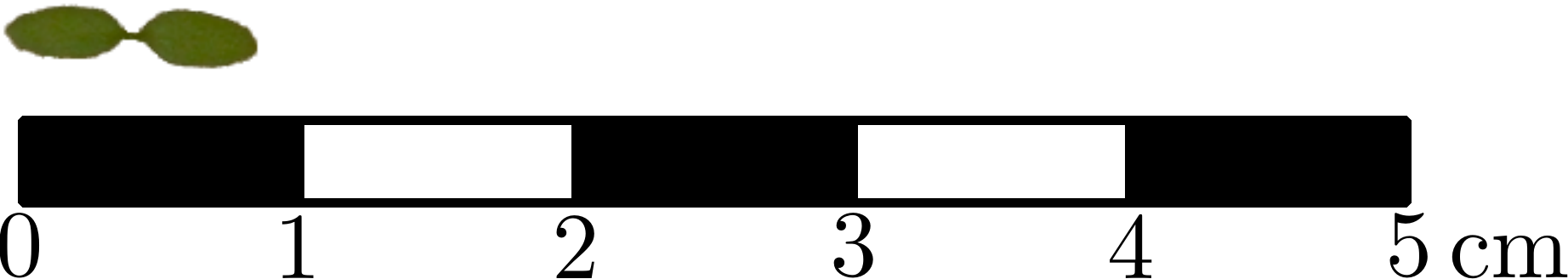}
	}
	\hfill
	\subfloat[Shepherd's Purse]{
		\includegraphics[width=0.3\textwidth]{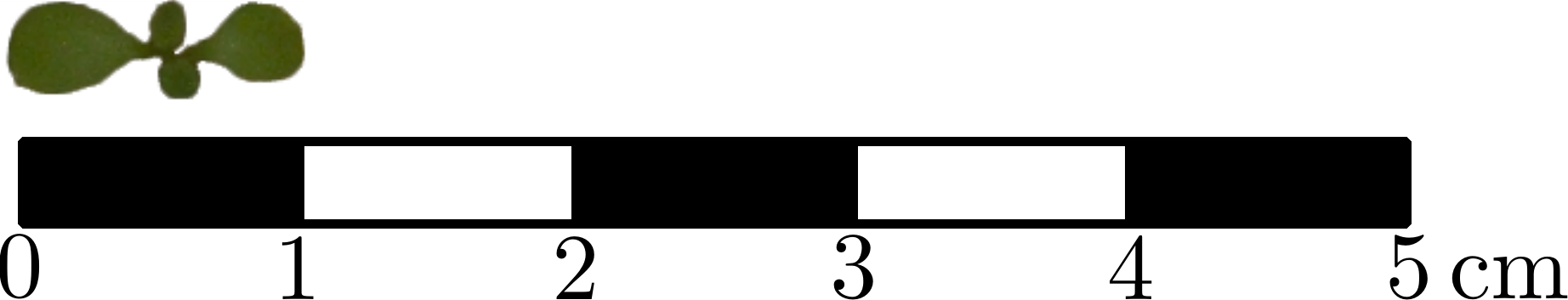}
	}
	
\caption{Sample images of species used for shape classification. Images are created by \cite{Dyrmann2013}.}
\end{figure}

\FloatBarrier


%% file: src/acknowledgement.tex
\section*{Acknowledgement}
The authors would like to give a special thanks to John Hallam, Maersk Institude, University of Southern Denmark for supervising and proof reading the manuscript. Part of the work have been performed under the project "Graduering af fungicider og herbicider i kartofler og korn", P-no. 1003407351.